\documentclass[sigconf]{acmart}

\usepackage{graphicx}
\usepackage{balance}
\usepackage{epsfig}
\usepackage{amssymb}
\usepackage{amsbsy}
\usepackage{amsmath}
\usepackage{amsfonts}

\def\BibTeX{{\rm B\kern-.05em{\sc i\kern-.025em b}\kern-.08emT\kern-.1667em\lower.7ex\hbox{E}\kern-.125emX}}
    
\copyrightyear{2018} 
\acmYear{2018} 
\setcopyright{acmcopyright}
\acmConference[DYNAMICS '18]{DYnamic and Novel Advances in Machine Learning and Intelligent Cyber Security Workshop}{December 3--4, 2018}{San Juan, Puerto Rico, USA}
\acmBooktitle{DYnamic and Novel Advances in Machine Learning and Intelligent Cyber Security Workshop (DYNAMICS '18), December 3--4, 2018, San Juan, Puerto Rico, USA}
\acmPrice{15.00}
\acmDOI{10.1145/3306195.3306196}
\acmISBN{978-1-4503-6218-4/18/12}

\begin{document}
\title{Crafting Adversarial Examples For Speech Paralinguistics Applications}

\author{Yuan Gong}
\affiliation{%
\institution{Department of Computer Science and Engineering} 
\institution{University of Notre Dame}
}
\email{ygong1@nd.edu}

\author{Christian Poellabauer}
\affiliation{%
\institution{Department of Computer Science and Engineering} 
\institution{University of Notre Dame}
}
\email{cpoellab@nd.edu}

\begin{abstract}

Computational paralinguistic analysis is increasingly being used in a wide range of cyber applications, including security-sensitive applications such as speaker verification, deceptive speech detection, and medical diagnostics. While state-of-the-art machine learning techniques, such as deep neural networks, can provide robust and accurate speech analysis, they are susceptible to adversarial attacks. In this work, we propose an end-to-end scheme to generate adversarial examples for computational paralinguistic applications by perturbing directly the raw waveform of an audio recording rather than specific acoustic features. Our experiments show that the proposed adversarial perturbation can lead to a significant performance drop of state-of-the-art deep neural networks, while only minimally impairing the audio quality.

\end{abstract}

\keywords{Paralinguistics, adversarial examples, speech processing, computer security}

\maketitle

\section{Introduction}
\subsection{Background}
\label{sec:background}

Computational speech paralinguistic analysis is rapidly turning into a mainstream topic in the field of artificial intelligence. In contrast to computational linguistics, computational paralinguistics analyzes how people speak rather than what people say~\cite{schuller2013computational}. Studies have shown that human speech not only contains the basic verbal message, but also paralinguistic information, which can be used (when combined with machine learning) in a wide range of applications, such as speaker verification~\cite{heigold2016end,lei2014novel,variani2014deep}, speech emotion recognition~\cite{trigeorgis2016adieu,gong2017continuous}, conversation analysis~\cite{lee2011analysis,laskowski2008modeling}, speech deception detection~\cite{schuller2016interspeech}, and medical diagnostics~\cite{schoentgen2006vocal,bocklet2011detection,daudet2017portable,malyska2005automatic}. Many of these applications are cloud-based, security sensitive, and must be very reliable, e.g., speaker verification systems used to prevent unauthorized access should have a low false positive rate, while systems used to detect deceptive speech should have a low false negative rate. A threat to these types of systems, which has not yet found widespread attention in the speech paralinguistic research community, is that the machine learning models these systems rely on can be vulnerable to adversarial attacks, even when these models are otherwise very robust to noise and variance in the input data. That is, such systems may fail even with very small, but well-designed perturbations of the input speech, leading the machine learning models to produce wrong results.

\begin{figure}[t]
  \centering
  \includegraphics[width=6cm]{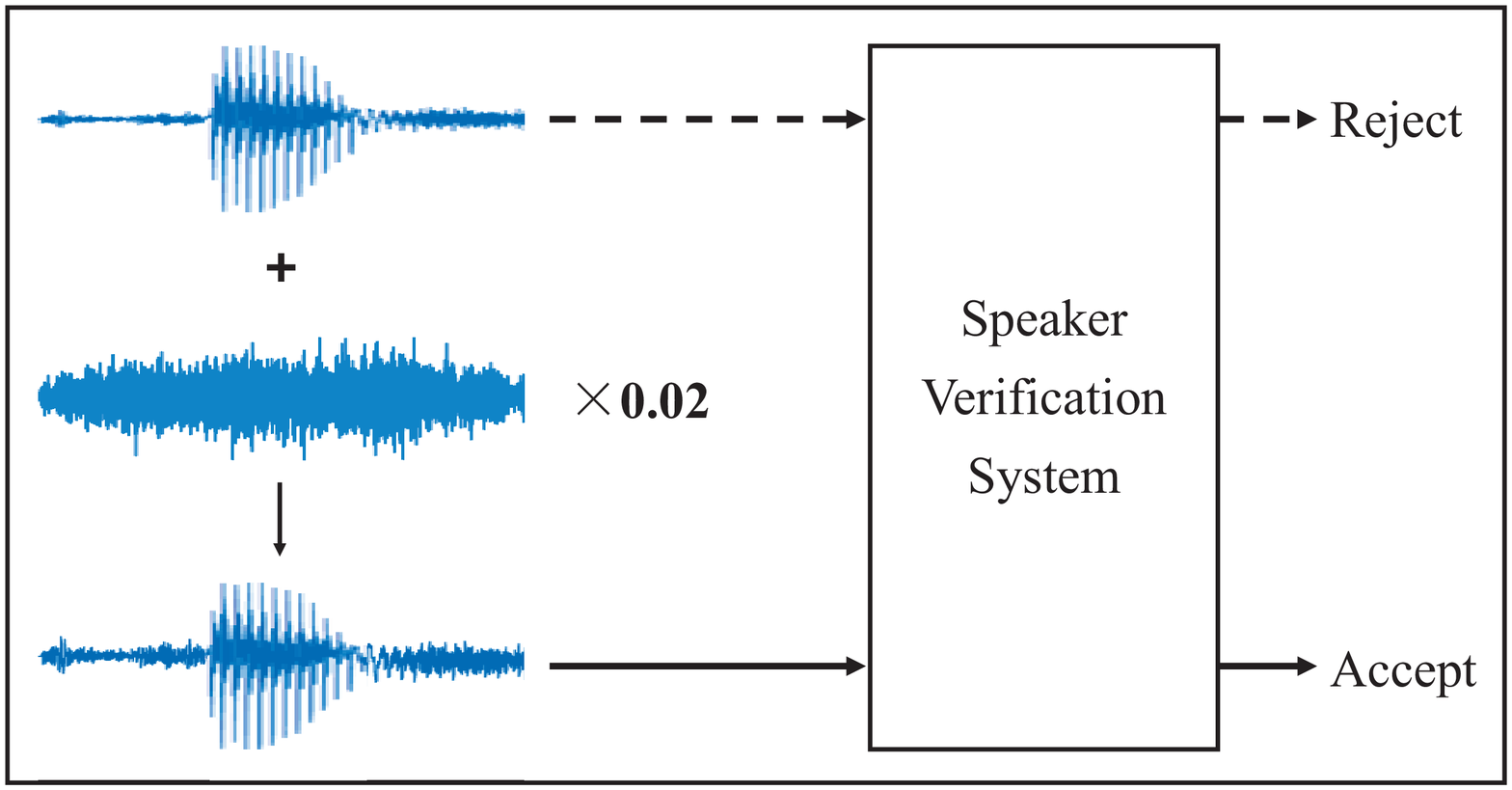}
  \caption{Illustration of an attack on a speaker verification system: the attacker can make the system accept an illegal input by adding a well-designed small perturbation. }
  \label{fig:ilus}
\end{figure}

An adversarial attack can lead to severe security concerns in security-sensitive computational paralinguistic applications. Since it is difficult for humans to distinguish adversarial examples generated by an attacking algorithm from the original legitimate speech samples, it is possible for an attacker to impact speech-based authentication systems, lie detectors, and diagnostic systems (e.g., Figure~\ref{fig:ilus} illustrates an example of attacking a speaker verification system, which could allow a burglar to enter a house by deceiving a voice-based smart lock). In addition, with rapidly growing speech databases (e.g., recordings collected from speech-based IoT systems), machine learning techniques are increasingly being used to extract hidden information from these databases. However, if these databases are polluted with adversarial examples, the conclusions drawn from the analysis can be false or misleading. 

As a consequence, obtaining a deeper understanding of this problem will be essential to learning how to prevent future adversarial attacks from impacting speech analysis. Toward this end, this paper proposes a method to generate adversarial speech examples from original examples for speech paralinguistics applications. Specifically, instead of manipulating specific speech acoustic features, our goal is to perturb the raw waveform of an audio directly. Our experiments show that the resulting adversarial speech examples are able to impact the outcomes of state-of-the-art paralinguistic models, while affecting the sound quality only minimally.

\subsection{Related Work} \label{sec:relatedWork}
\subsubsection{Computer Vision}
The vulnerability of neural networks to adversarial examples, particularly in the field of computer vision, was first discovered and discussed in~\cite{szegedy2013intriguing}. In~\cite{goodfellow2014explaining}, the authors analyzed the reasons and principles behind the vulnerability of neural networks to adversarial examples and proposed a simple yet effective gradient based attack approach, which has been widely used in later studies. In recent years, properties of adversarial attacks have been studied extensively, e.g., in~\cite{kurakin2016adversarial}, the authors found that machine learning models may make wrong predictions even when using images that are fed to the model via a camera instead of directly applying the adversarial examples as input. Further, in~\cite{papernot2016practical}, the authors found that it is possible to attack a machine learning model even without knowing the details of the model. 

\subsubsection{Speech and Audio Processing} 
In~\cite{carlini2016hidden,02vaidya2015cocaine}, the authors proposed an approach to generate hidden voice commands, which can be used to attack speech recognition systems. These hidden voice commands are reconstructed from Mel-frequency cepstral coefficients (MFCC). They are unrecognizable by the human ear and not similar to any legitimate speech. Hence, strictly speaking, they are not machine learning adversarial examples. In~\cite{kereliuk2015deep}, the authors proposed an approach to generate adversarial examples of music by applying perturbations on the magnitude spectrogram. However, rebuilding time-domain signals from magnitude spectrograms is difficult, because of the overlapping windows used for analysis, which makes adjacent components in the spectrogram dependent. Similarly, in~\cite{06cisse2017houdini}, the authors add perturbations to the magnitude spectrograms and make the adversarial examples fail to be recognized by both known or unknown automatic speech recognition systems. In~\cite{itergenerating}, the authors also proposed an approach to generate adversarial speech examples, which can mislead speech recognition systems. They first extract MFCC features from the original speech, add perturbations to the MFCC features, and then rebuild speech from the perturbed MFCC features. While the rebuilt speech samples are still recognizable by the human ear, they are very different from the original samples, because extracting MFCC from audio is a lossy conversion. In~\cite{carlini2018audio,taori2018targeted,alzantot2018did}, the authors proposed targeted attacks on deep neural network based speech recognition system.

To the best of our knowledge, all prior speech-related efforts aim to attack speech recognition (i.e., speech to text) systems. Speech recognition (typically a transcription task) is very different from computational paralinguistic tasks (which are typically classification or regression tasks). As mentioned in Section~\ref{sec:background}, adversarial attacks on computational paralinguistic applications can also lead to severe security concerns, but they have not been studied before. Therefore, it will be interesting to study if adversarial attacks are also effective in computational paralinguistics applications. Further, most prior efforts add perturbations at the {\em feature level} and require a {\em reconstruction step} using acoustic and spectrogram features, which will further modify and impact the original data (in addition to the actual perturbation). This additional modification can make adversarial examples sound strange to the human ear (i.e., they become recognizable). In comparison, in the computer vision field, adversarial examples are perturbed directly on the pixel value level and only contain very minor amounts of noise that are difficult or even impossible to detect by the human eye. Therefore, in order to avoid the downsides of performing the inverse conversions from features to audio, in this work we propose an end-to-end approach to crafting adversarial examples by directly modifying the original waveforms.

\subsection{Contributions of this Paper}

\begin{itemize}
\item To the best of our knowledge, this is the first work on adversarial attacks in the field of computational speech paralinguistics. We expect that our results and discussions will bring useful insights for future studies and efforts in building more robust machine learning models.

\item We propose an end-to-end speech adversarial example generation scheme
that does not require an audio reconstruction step. The perturbation is added directly to the raw waveform rather than the acoustic features or spectrogram features so that no lossy conversion from the features back to the waveform is needed.

\item We describe the vanishing gradients problem when using a gradient based attack approach on recurrent neural networks. We address this problem by using a substitution network that replaces the recurrent structure with a feed-forward convolutional structure. We believe that this solution is not limited to our application, but can be used in other applications with sequences of inputs.
\item We provide comprehensive experiments with three different speech paralinguistics tasks, which empirically prove the effectiveness of the proposed approach. The experimental results also indicate that the adversarial examples can be generalized to other models.

\end{itemize}

\begin{figure}[h]
  \centering
  \includegraphics[width=5.0cm]{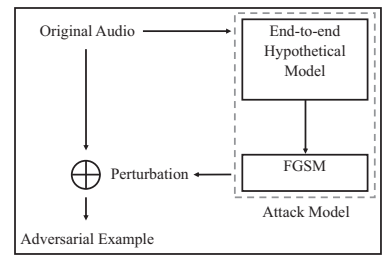}
  \caption{The proposed end-to-end adversarial attack scheme.}
  \label{fig:e2e}
\end{figure}

\section{Speech Adversarial Examples}

The block diagram of the proposed speech adversarial attack scheme is shown in Figure~\ref{fig:e2e}. In this section, we first provide a formal definition of the adversarial example generation problem. We then briefly review the gradient based approach used in this work and discuss its use in our application. We explain the limitations of acoustic-feature-level adversarial attacks and why building an end-to-end speech adversarial attack scheme is essential. Further, we discuss the selection of a hypothetical model for the attack. Finally, we describe the vanishing gradient problem in gradient based attacks on recurrent neural networks and address this problem using a model substitution approach. 

\subsection{Definitions}
We describe a classifier that maps raw speech audio waveform vectors into a discrete label set as $f:\mathbb{R}^n\rightarrow\{1...k\}$ and the parameters of $f$ as $\mathbf{\theta}$. We further assume that there exists a continuous loss function associated with $f$ described as $J:\mathbb{R}^n\times\{1...k\}\rightarrow\mathbb{R}^+$. We describe a given speech audio waveform using $\mathbf{x}\in\mathbb{R}^n$, the ground-truth label of $\mathbf{x}$ using $y$, and a perturbation on the vector using $\mathbf{\eta}\in\mathbb{R}^n$. Our goal is to address the following optimization problem: 
\begin{equation}
\label{optimization}
\text{Minimize} \left \| \mathbf{\eta}\right \|\quad \text{s.t.}\  f(\mathbf{x}+\mathbf{\eta})\neq f(\mathbf{x})
\end{equation}

\subsection{Gradient Based Adversarial Attacks}
Due to the non-convexity of many machine learning models (e.g., deep neural networks), the exact solution of the optimization in Equation~\ref{optimization} can be difficult to obtain. Therefore, gradient based methods~\cite{goodfellow2014explaining} have been proposed to find an approximation of such an optimization problem. The gradient based method under the max-norm constraint is referred to as \emph{Fast Gradient Sign Method} (FGSM). The FGSM is then used to generate a perturbation $\mathbf{\eta}$:
\begin{equation} \label{equ:FGSM}
\mathbf{\eta} = \epsilon\text{sign}(\nabla_{\mathbf{x}}J(\mathbf{\theta},\mathbf{x},y))
\end{equation}

Let $\mathbf{w}$ be the weight vector of a neuron. Consider the situation when the neuron takes the entire waveform $\mathbf{x}$ as input, then, when we add a perturbation $\mathbf{\eta}$, the activation of the neuron will be:
\begin{equation}
activation = \mathbf{w}^\mathrm{T}(\mathbf{x}+\mathbf{\eta}) = \mathbf{w}^\mathrm{T}\mathbf{x}+\mathbf{w}^\mathrm{T}\mathbf{\eta}
\end{equation}

The change of the activation is then expressed as:
\begin{equation}\label{equ:activation}
\begin{split}
\Delta\ activation & = \mathbf{w}^\mathrm{T}\mathbf{\eta} \\
& = \epsilon \mathbf{w}^\mathrm{T}\text{sign}(\nabla_{\mathbf{x}}J(\mathbf{w},\mathbf{x},y))\\
& = \epsilon \mathbf{w}^\mathrm{T}\text{sign}(\mathbf{w})
\end{split}
\end{equation}

Assume that the average magnitude of $\mathbf{w}$ is expressed as $m$, then note that $\mathbf{w}$ has the same dimension with $\mathbf{x}$ of $n$:
\begin{equation}
\Delta\ activation = \epsilon m n
\end{equation}

Since $\left \| \mathbf{\eta}\right \|_\infty$ does not grow with $n$, we find that even when $\left \| \mathbf{\eta}\right \|_\infty$ is fixed, the change of the activation of the neuron grows linearly with the dimensionality $n$, which indicates that for a high dimensional problem, even a small change in each dimension can lead to a large change of the activation, which in turn will lead to a change of the output, because even for non-linear neural networks, the activation function primarily operates linearly in the non-saturated region. We call this the \emph{accumulation effect}. Speech data processing, when done in an end-to-end manner, is an extremely high dimensional problem. Using typical sampling rates of 8kHz, 16kHz, or 44.1kHz, the dimensionality of the problem can easily grow into the millions for a 30-second audio. Thus, taking the advantage of the accumulation effect, a small average well-designed perturbation for each data point can cause a large shift in the output decision.

\begin{figure}[h]
  \centering
  \includegraphics[width=5.9cm]{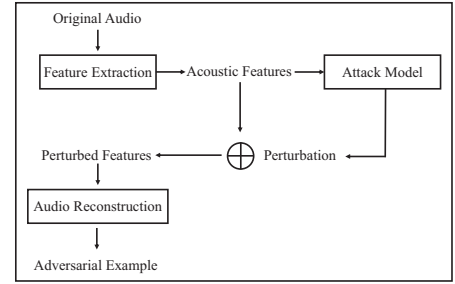}
  \caption{A feature-level attack scheme.}
  \label{fig:feature}
\end{figure}

\subsection{End-to-end Adversarial Example Generation}

As discussed in the related work section, most previous approaches to audio adversarial example generation are on the acoustic feature level. As shown in Figure~\ref{fig:feature}, acoustic features are first extracted from the original audio, the perturbation is added to the acoustic features, and then audio is reconstructed from the perturbed acoustic features. In recent years, end-to-end machine learning models that do not explicitly rely on hand-crafted acoustic features have become a mainstream technology. When targeting these models, adversarial example generation schemes using acoustic features might still work (because end-to-end models may implicitly rely on such acoustic features), but they are less efficient due to the following reasons:

\begin{itemize}
    \item Acoustic feature extraction is usually a lossy conversion and reconstructing audio from acoustic features cannot recover this loss. For example, in \cite{itergenerating}, an adversarial attack is conducted on the MFCC features, which typically represent each audio frame of 20-40ms (160-320 data points if the sample rate is 8kHz) using a vector of 13-39 dimensions. The conversion loss can be significant and even be larger than the adversarial perturbation due to the information compression. Thus, the final perturbation on the audio waveform consists of the adversarial perturbation plus an extra perturbation caused by the lossy conversion:
    \begin{equation}
    \mathbf{\eta}_{\text{overall}} = \mathbf{\eta}_{\text{conversion\ loss}} + \mathbf{\eta}_{\text{adversarial}}
    \end{equation}
    
   \item Adversarial attacks on acoustic features and on raw audio waveform are actually two different optimization problems:
   \begin{equation}
       \text{Minimize} \left \| \mathbf{\eta}_{\text{acoustic\ feature}}\right  \| \neq \text{Minimize} \left \| \mathbf{\eta}_{\text{audio}}\right \|
   \end{equation}
   Since most acoustic features do not have a linear relationship with the audio amplitude, it is possible that a small perturbation on acoustic features will lead to large perturbations on the audio waveform and vice versa. Thus, an adversarial attack on acoustic features might not even be an approximation of an adversarial attack on the raw audio.
\end{itemize}

\begin{figure}[t]
  \centering
  \includegraphics[width = 0.72\linewidth]{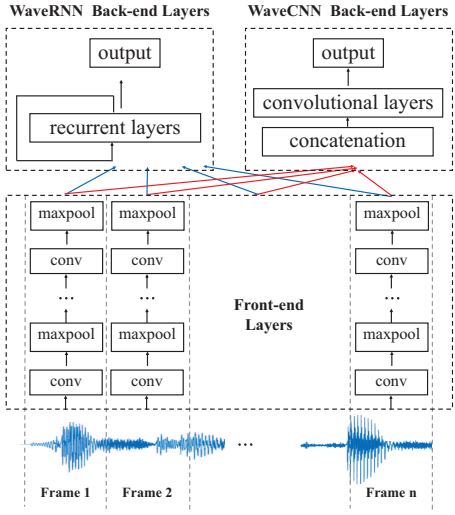}
  \caption{Network topology of WaveRNN and WaveCNN.}
  \label{fig:topo}
\end{figure}

In order to overcome the above-mentioned limitations, we propose an {\em end-to-end adversarial attack scheme} that directly perturbs the raw audio waveform. Compared to the feature-level attack scheme, the proposed scheme completely abandons the feature extraction and audio reconstruction steps to avoid any possible conversion loss. The optimization is then directly targeted at the perturbation on the raw audio.
A key component of the proposed scheme is to use an end-to-end machine learning model that is able to directly map the raw audio to the output as our hypothetical model. A good choice of such a hypothetical model is the model proposed in~\cite{trigeorgis2016adieu}; we refer to this model as WaveRNN in the remainder of this paper. WaveRNN is the first network that learns directly from raw audio waveforms to paralinguistic labels using recurrent neural networks (RNNs) and has become a state-of-the-art baseline approach for multiple paralinguistic applications due to its excellent performance compared to previous models~\cite{schuller2017interspeech}. As shown in Figure~\ref{fig:topo}, WaveRNN first segments the input audio sequence into frames of 40ms and processes each frame separately in the first few layers (front-end layers). The output of the front-end layers is then fed, in order of time, to the back-end recurrent layers. 

Previous studies have shown that adversarial examples are able to generalize, i.e., examples generated for attacking the hypothetical model are then often able to attack other models, even when they have different architectures~\cite{goodfellow2014explaining}. Further, the more similar the structures of the hypothetical model and a practical model are, the better the practical attack performance will be. Thus, the adversarial examples generated for the WaveRNN model are expected to have good attack performance with a variety of state-of-the-art WaveRNN-like networks widely used in different paralinguistic tasks. However, in our work, we also discovered the {\em vanishing gradient problem}, which prevents us from using WaveRNN directly as the hypothetical model. This problem is discussed in the next section.

\subsection{The Vanishing Gradient Problem of Gradient Based Attacks on RNNs}

One basic prerequisite of gradient based attacks is that the required gradients $\mathbf{g} = \nabla_{\mathbf{x}}J(\mathbf{\theta},\mathbf{x},y)$ can be computed efficiently~\cite{goodfellow2014explaining}. While, in theory, the gradient based method applies as long as the model is differentiable~\cite{papernot2016crafting}, even if there are recurrent connections in the model, we observe that when we use RNNs to process long input sequences (such as speech signals), most elements in $\mathbf{g}$ go to zero except the last few elements. This means that the gradient of the early input in the sequence is not calculated effectively, which will cause the perturbation $\mathbf{\eta} = \epsilon\text{sign}(\nabla_{\mathbf{x}}J(\mathbf{\theta},\mathbf{x},y))$ to only have meaningful values in the last few elements. In other words, the perturbation is only correctly added to the end of the input sequence. Since this problem has not been reported before (calculating the gradients with respect to the input $\nabla_{\mathbf{x}}J(\mathbf{\theta},\mathbf{x},y)$ is not typical, while $\nabla_{\mathbf{\theta}}J(\mathbf{\theta},\mathbf{x},y)$ is), we formalize and describe it mathematically in the following paragraphs.

We assume one single output (rather than a sequence of outputs) for each input sequence, which is the typical case for paralinguistic applications. Consider the simplest one-layer RNN and let $\mathbf{s}_t=tanh(\mathbf{U}\mathbf{x}_t+\mathbf{W}\mathbf{s}_{t-1})$ be the state of neurons at time step $t$, $y$ be the ground-truth label, $\hat{y}$ be the prediction, $J(y,\hat{y})$ be the loss function, and $n$ be the number of time steps of the sequence. The vanishing gradient problem in gradient based attacks can be described as follows:
\begin{equation}\label{equ:vanish}
\frac{\partial J}{\partial \mathbf{x}_i} = \frac{\partial J}{\partial \hat{y}} \frac{\partial \hat{y}}{\partial \mathbf{s}_n} \frac{\partial \mathbf{s}_n}{\partial \mathbf{s}_{i}} \frac{\partial \mathbf{s}_{i}}{\partial \mathbf{x}_i }
\end{equation}
We can use the chain rule to expand $\frac{\partial \mathbf{s}_n}{\partial \mathbf{s}_{i}}$:
\begin{equation}\label{equ:conMul1}
\begin{split}
    \frac{\partial \mathbf{s}_n}{\partial \mathbf{s}_{i}} & = \frac{\partial \mathbf{s}_n}{\partial \mathbf{s}_{n-1}}\frac{\partial \mathbf{s}_{n-1}}{\partial \mathbf{s}_{n-2}}\cdots \frac{\partial \mathbf{s}_{i+1}}{\partial \mathbf{s}_{i}} \\
    &= \prod_{j=i}^{n-1}\frac{\partial \mathbf{s}_{j+1}}{\partial \mathbf{s}_{j}} \\ 
    &= \prod_{j=i}^{n-1}\frac{\partial tanh(\mathbf{U}\mathbf{x}_t+\mathbf{W}\mathbf{s}_{j})}{\partial \mathbf{s}_{j}}
\end{split}
\end{equation}

Since the derivative of the $tanh$ function is in $[0,1]$, we have:
\begin{equation} \label{equ:conMul}
0\leq \left \| \frac{\partial tanh(\mathbf{U}\mathbf{x}_t+\mathbf{W}\mathbf{s}_{j})}{\partial \mathbf{s}_{j}} \right \|_{2} \leq1 
\end{equation}

When the time interval between $i$ and $n$ becomes larger, especially when $(n-i)\rightarrow\infty$, the continuous multiplication in Equation~\ref{equ:conMul1} will approach zero:
\begin{equation} \label{equ:0}
 \lim_{(n-i) \to \infty}\left \|  \frac{\partial \mathbf{s}_n}{\partial \mathbf{s}_{i}} \right \|_{2} = 0
\end{equation}

We then substitute Equation~\ref{equ:0} into Equation~\ref{equ:vanish}:
\begin{equation} \label{equ:vanishCon}
    \lim_{(n-i) \to \infty}\frac{\partial J}{\partial \mathbf{x}_i} = 0
\end{equation}

\begin{figure}[h]
  \centering
  \includegraphics[width=7.0cm]{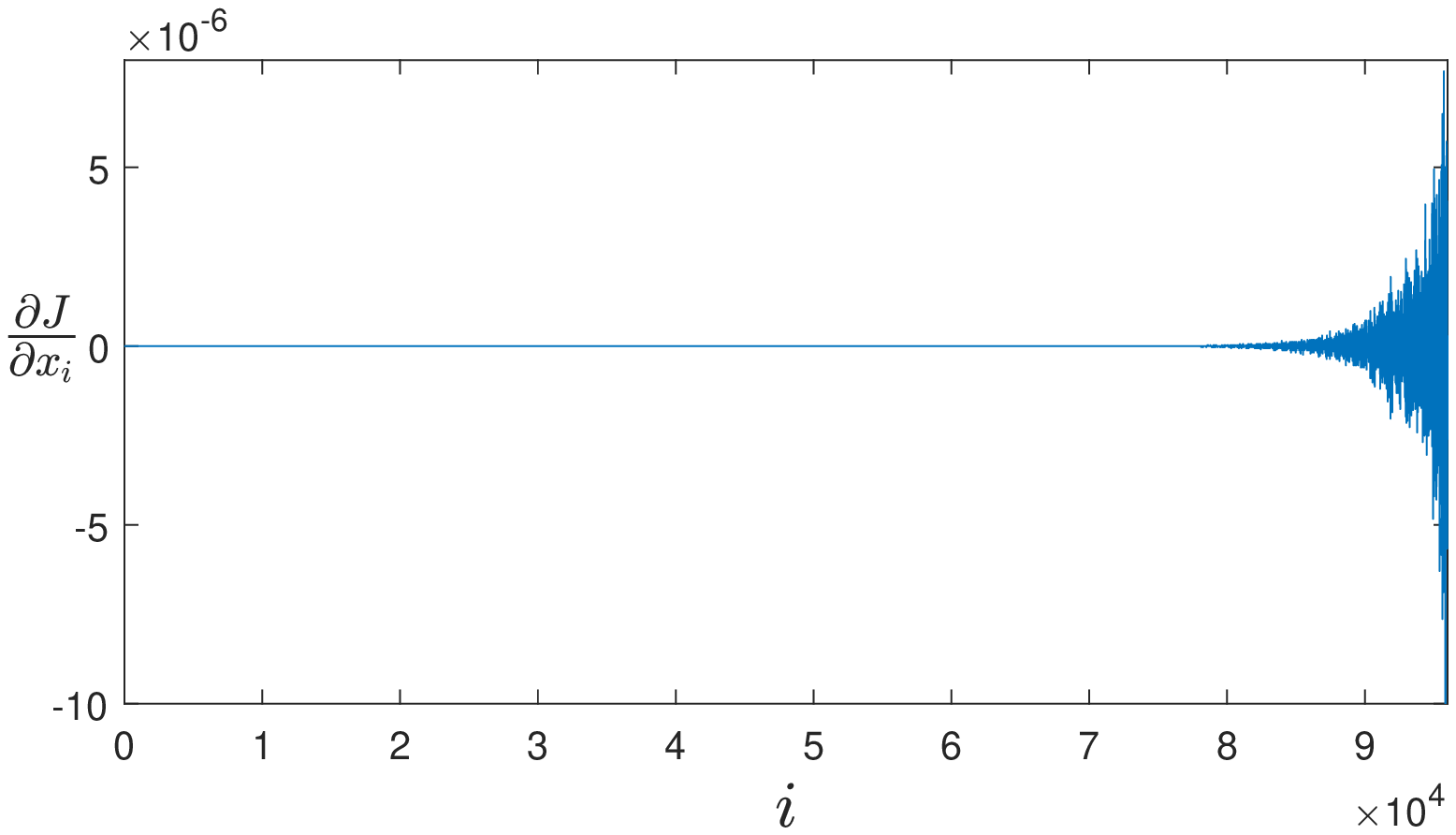}
  \includegraphics[width=7.0cm]{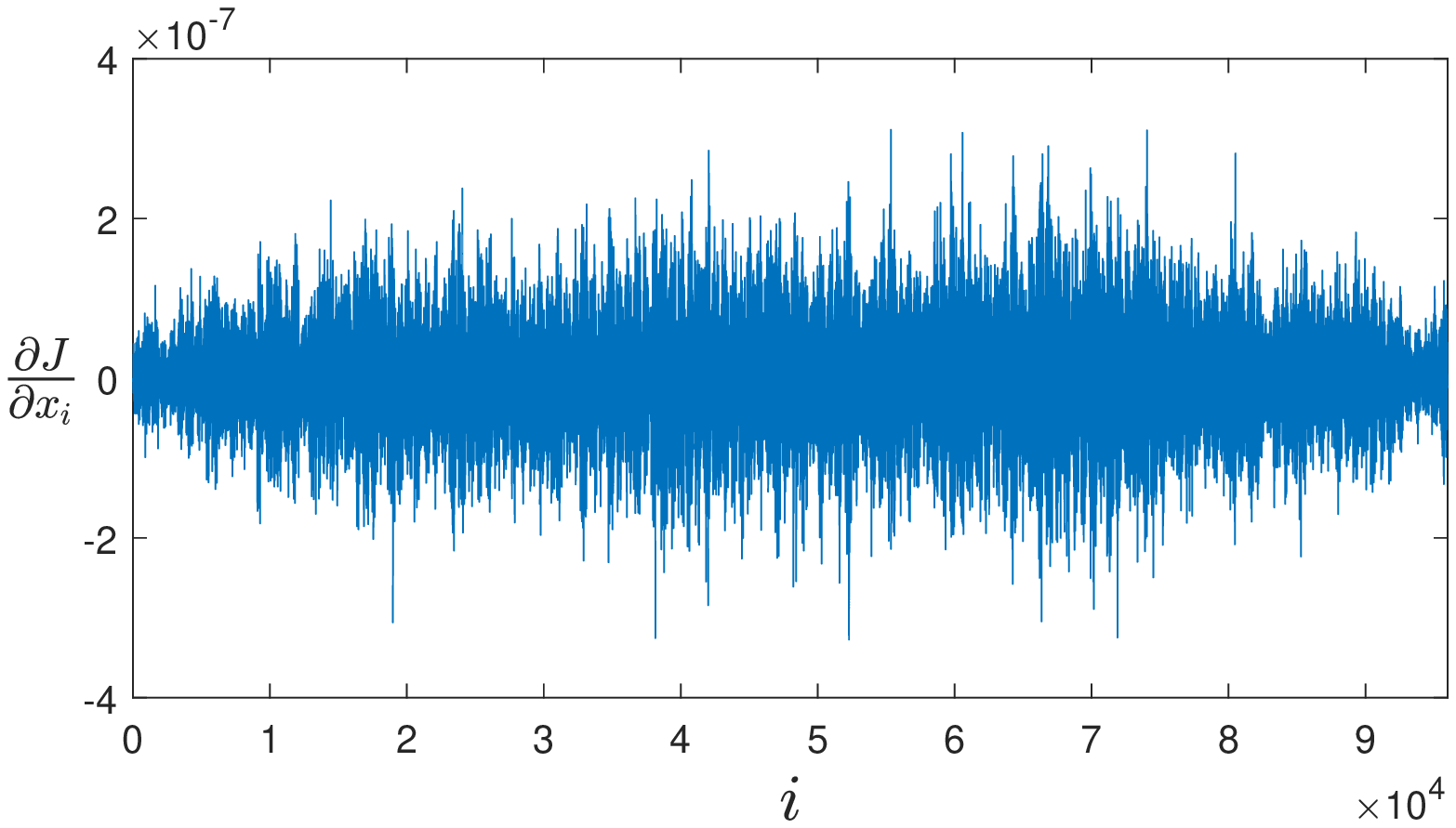}
  
  \caption{The vanishing gradient problem: partial derivatives of the RNN's loss function with respect to inputs at earlier time steps in a long sequence tend to disappear. Upper graph: the partial derivative of the loss function of WaveRNN (with LSTM layers) with respect to each input audio data point $\mathbf{x}_i$. Lower graph: the partial derivative of the loss function of WaveCNN with respect to each input audio data point $\mathbf{x}_i$.}
  \label{fig:vanish}
\end{figure}

Equation \ref{equ:vanishCon} indicates that for a long input sequence vector, partial derivatives of the loss function with respect to the inputs at earlier time steps are disappearing, which will make the gradient based method fail. We believe it is an inherent problem of RNNs. Interestingly, using Long Short Term Memory networks (LSTM) cannot fix the problem, potentially because LSTM will still forget some input information. Figure~\ref{fig:vanish} (upper graph) shows the partial derivative of the loss function of a trained WaveRNN model with respect to each input audio data point $\mathbf{x}_i$. Even when the WaveRNN uses the LSTM recurrent layers, the partial derivative of the first 60,000 data points are close to zero.

\subsection{The Substitution Model Approach}

According to the discussion in the last section, it is difficult to effectively calculate the gradient of the loss function with respect to the input sequence for RNNs. While RNNs are most commonly used in processing such kinds of sequence input problems, they are not indispensable. In order to fix the vanishing gradient problem, we propose a new network with a complete feedforward structure, which can be referred to as WaveCNN. Similar to WaveRNN, WaveCNN also first divides the input audio sequence into frames of 40ms and processes each frame separately in the front-end layers. But after that, instead of feeding the output of the front-end layers of each frame into recurrent layers, the WaveCNN approach concatenates the outputs of the front-end layers of all frames and feeds them to the following convolutional layers. The WaveCNN approach uses convolutional structures as a substitution for the recurrent structures. This modification eliminates the recurrent structures in the network and hence fixes the vanishing gradient problem. As shown in Figure~\ref{fig:vanish}, WaveCNN can calculate the gradient over all input data points effectively. The design of WaveCNN still retains the local receptive fields arithmetic of WaveRNN and the back-end convolutional layers are also able to capture temporal information. Actually, in our experiments, WaveCNN performs almost the same as WaveRNN for a series of tasks. We therefore use WaveCNN as the hypothetical model in our work.


\begin{figure*}[h]
  \centering
  \includegraphics[width=0.28\linewidth]{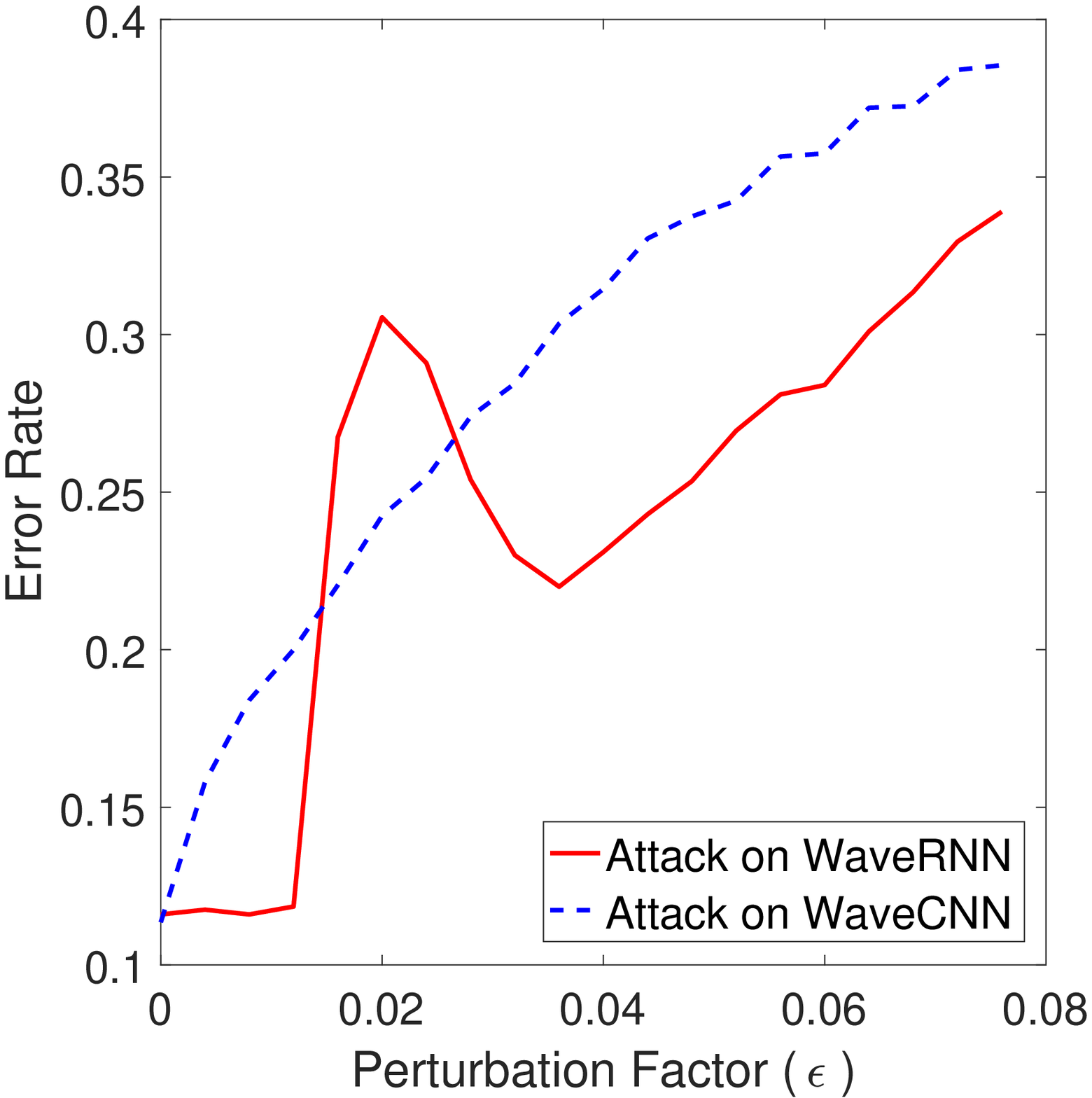}\hfill
  \includegraphics[width=0.28\linewidth]{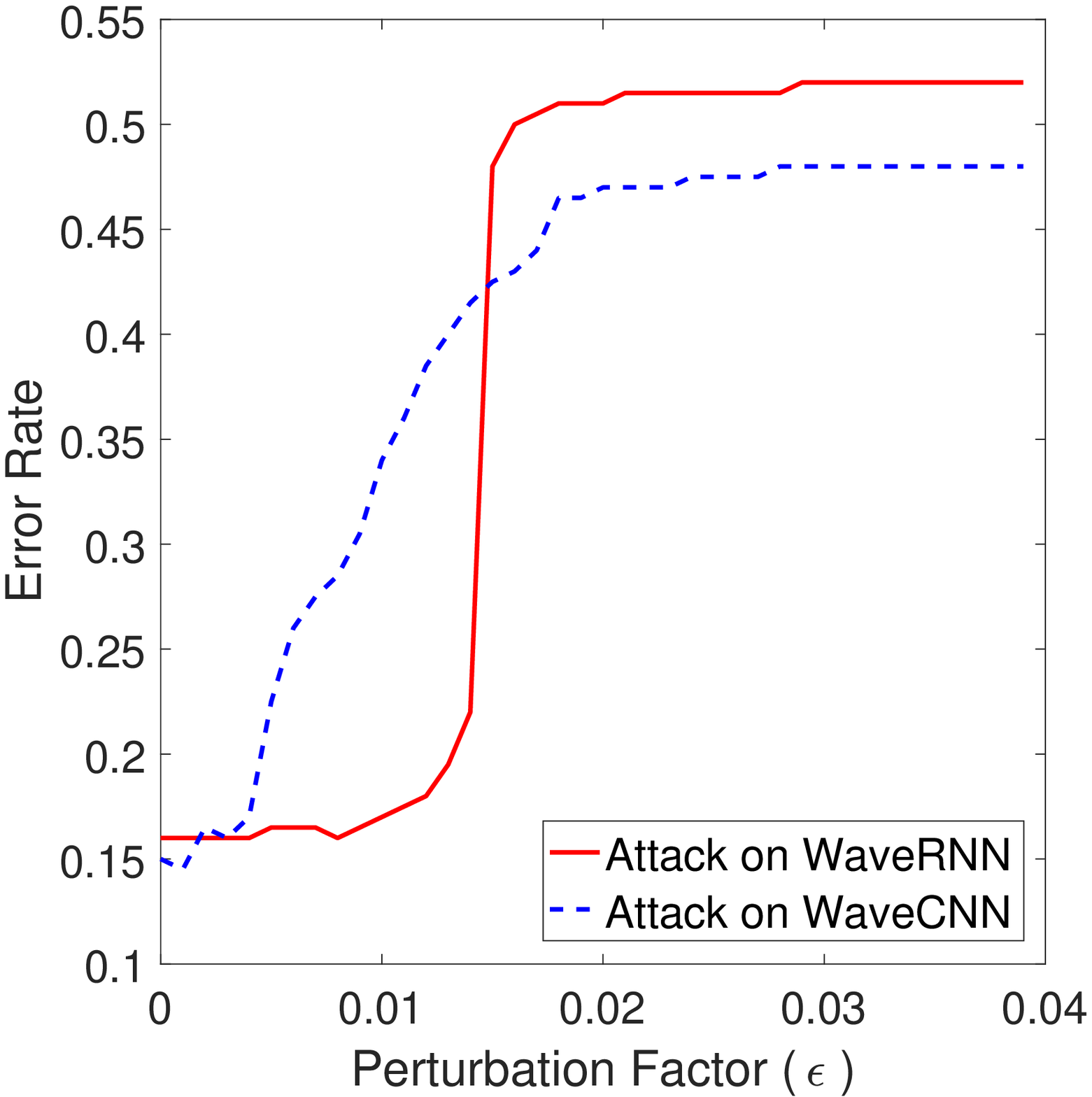}\hfill
  \includegraphics[width=0.28\linewidth]{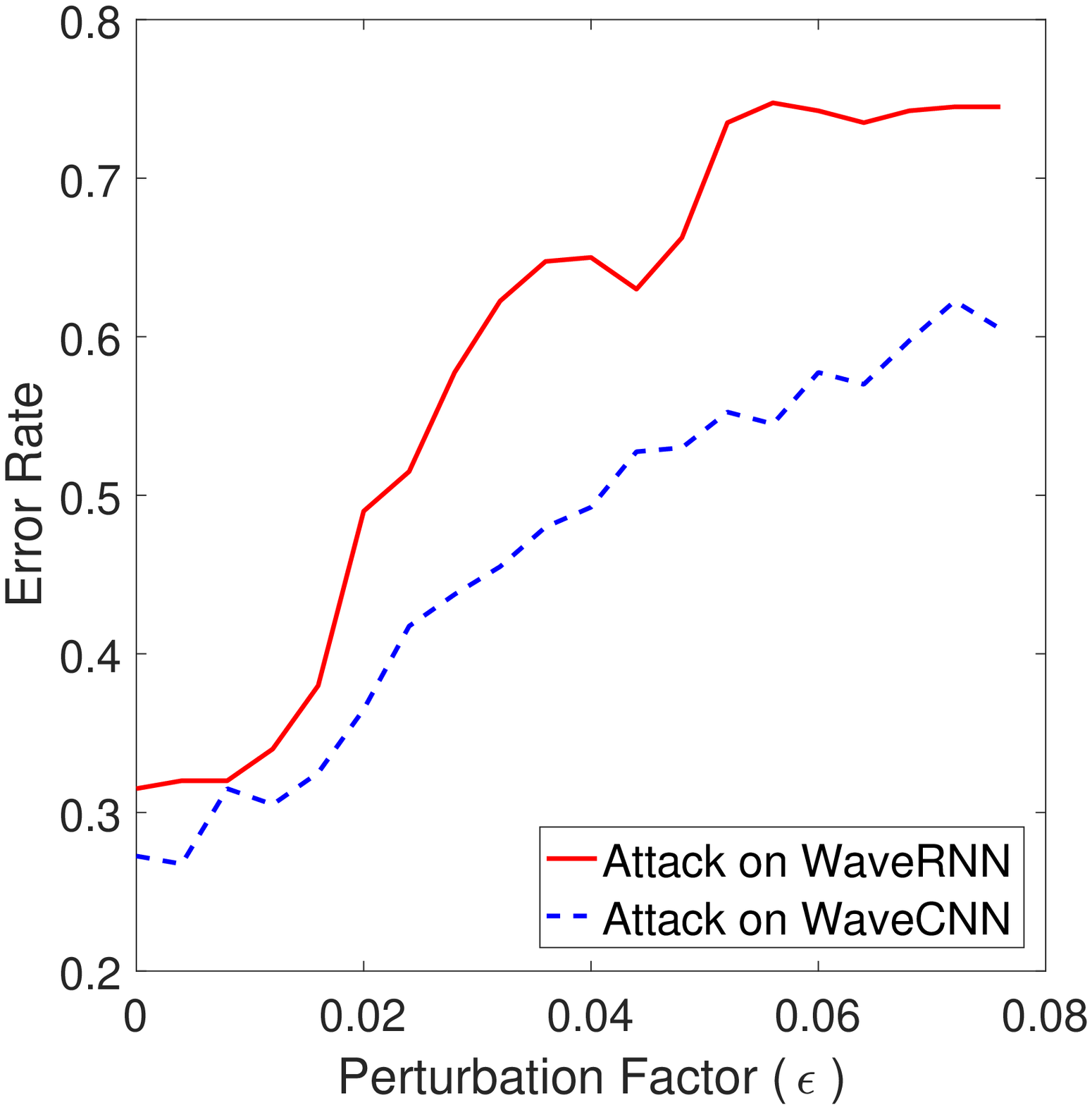}
  \caption{The error rate with different perturbation factors for the gender recognition task (left), emotion recognition task (middle), and speaker recognition task (right).}
  \label{fig:attackResult}
\end{figure*}

\begin{figure*}[h]
  \centering
  \includegraphics[width=0.36\linewidth]{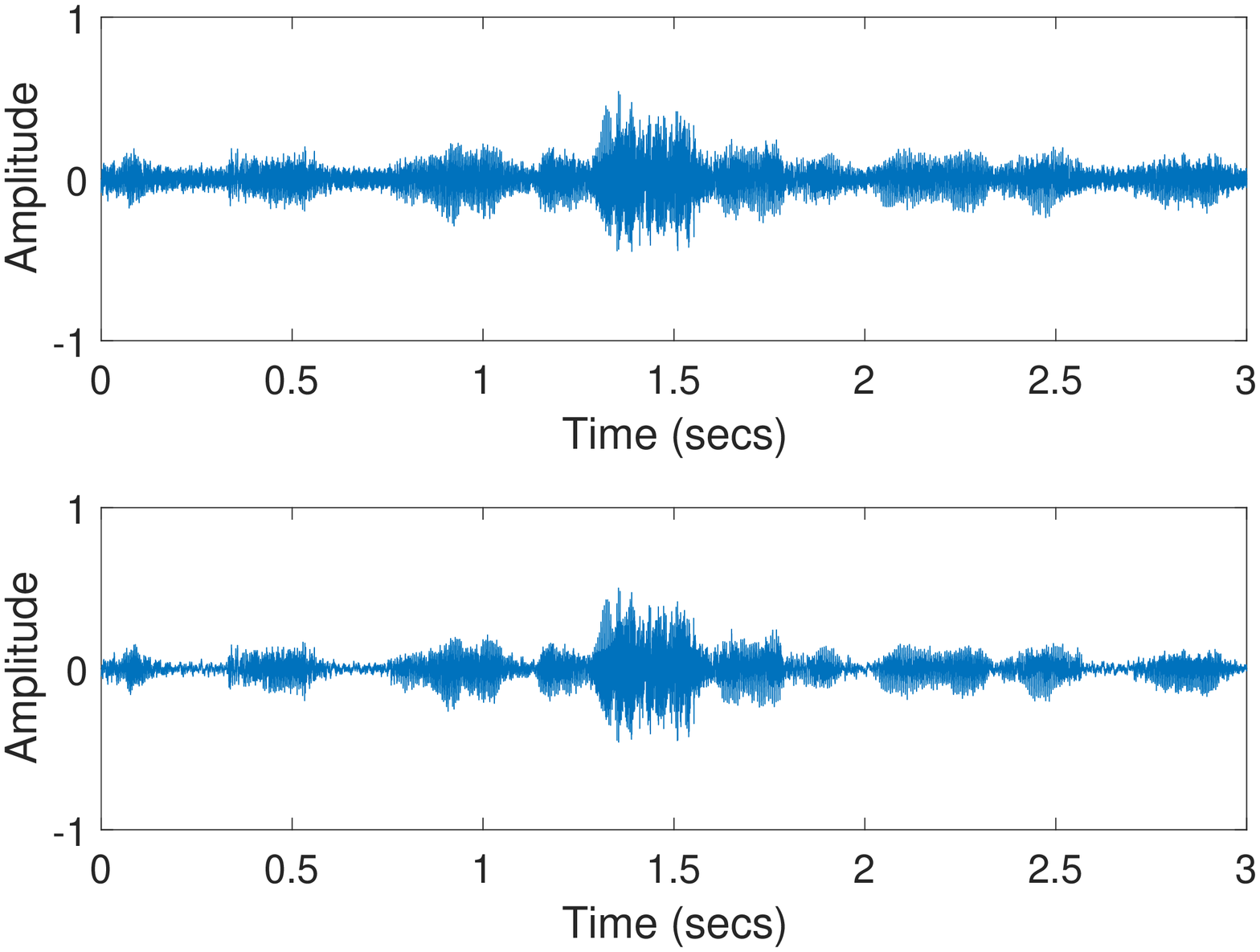}\hfill
  \includegraphics[width=0.378\linewidth]{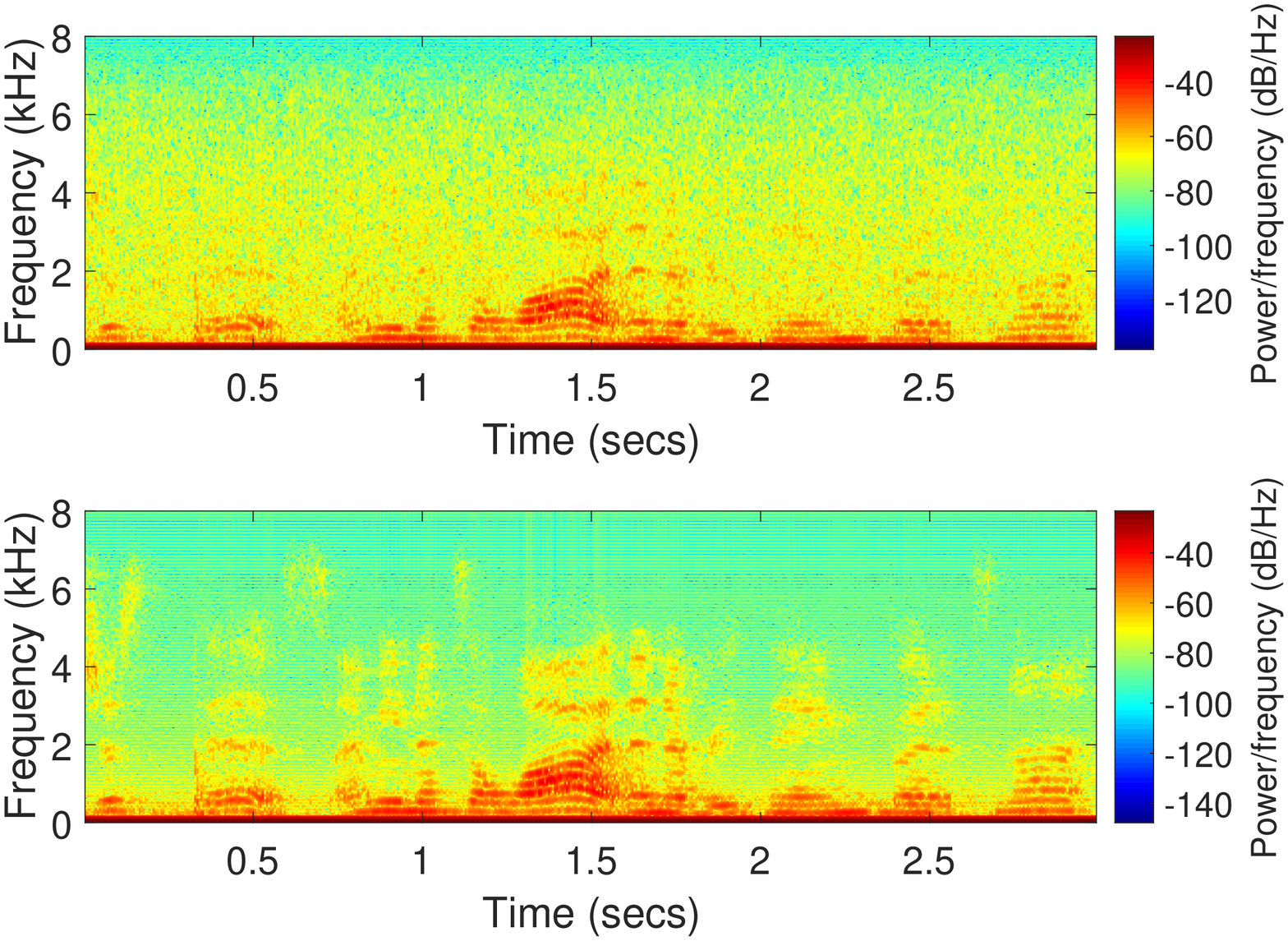}\hfill
   \includegraphics[width=0.216\linewidth]{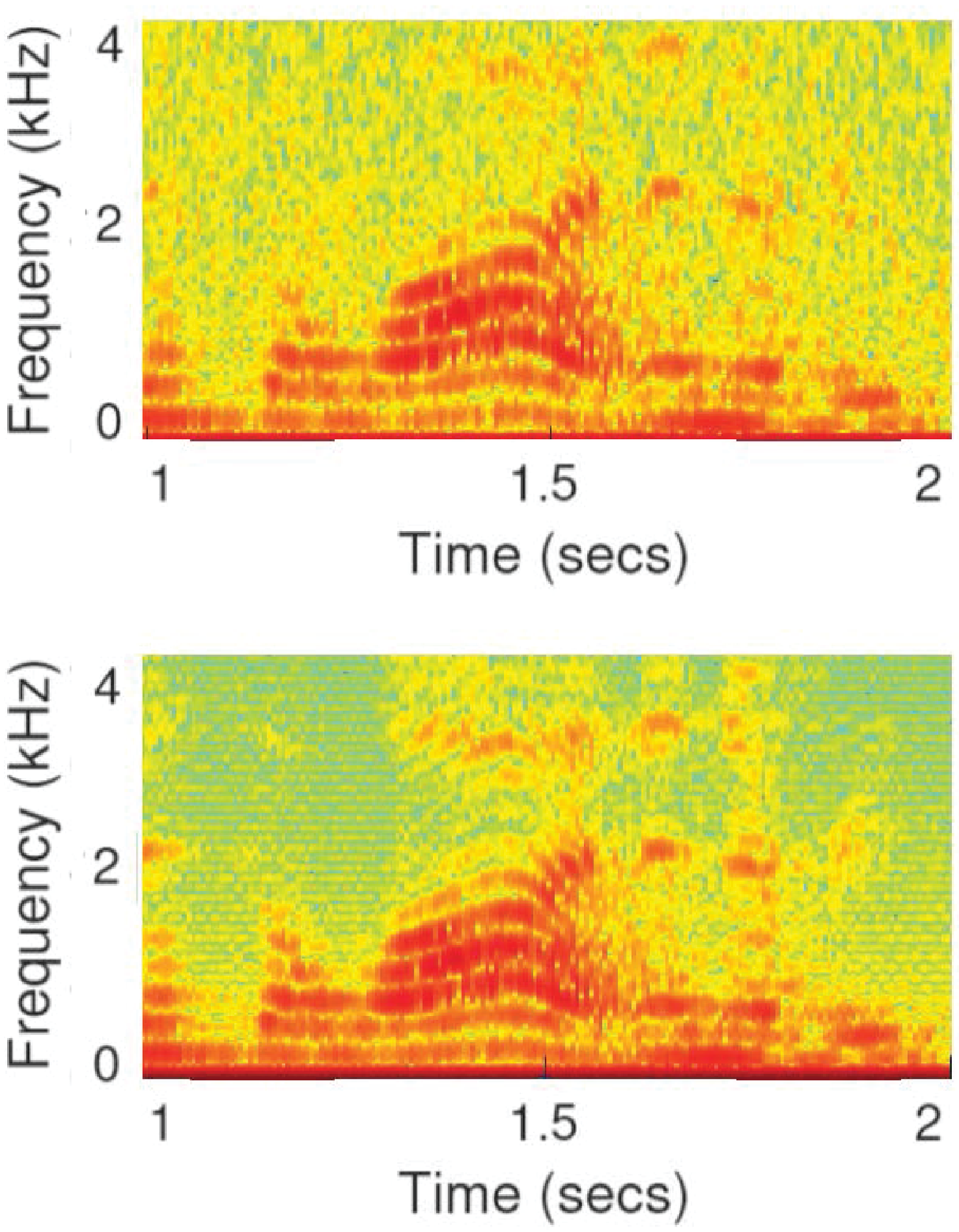}
  \caption{Comparison of the waveform (left), spectrogram (middle), and enlarged vocal spectrogram  (right) between the adversarial example with $\epsilon = 0.02$ (upper graphs) and original example (lower graphs).}
  \label{fig:perturbation}
\end{figure*}

\section{Experimental Evaluation}
\subsection{Dataset and Test Strategy}
We use the audio part of the IEMOCAP dataset \cite{busso2008iemocap} for our experiments, which is a commonly used database in speech paralinguistic research. The IEMOCAP database consists of 10,039 utterances (average length: 4.46s) of 10 speakers (5 male, 5 female). In order to make our experiments without loss of generality, we perform the experiments with three different speech paralinguistic tasks using IEMOCAP:

\begin{itemize}
\item \textbf{Gender Recognition:} A binary classification task of predicting the gender of the speaker. The IEMOCAP database consists of 5239 utterances from male speakers and 4800 utterances from female speakers.
\item \textbf{Emotion Recognition:} A binary classification task to distinguish sad speech from angry speech. The IEMOCAP database consists of 1103 utterances annotated as angry and 1084 utterances annotated as sad.
\item \textbf{Speaker Recognition:} A four-class classification task of predicting the identity of the speaker. We use the data of four speakers in IEMOCAP sessions 1\&2, which consist of two male speakers and two female speakers. The number of utterances of each speaker are 946, 873, 952, and 859, respectively.
\end{itemize}


Note that we simplified the emotion recognition task and the speaker recognition task in our experiment by limiting them to a two-class and four-class problem, respectively. This simplification makes the model have a better performance before being attacked. It is more meaningful to attack a model that originally has good performance. The simplification also makes the class distribution balanced in our experiments, therefore, accuracy is an effective metric for evaluating the performance of the model. All experiments are conducted in a hold-out test strategy, i.e., 75\%, 5\%, and 20\% of the data is used for training, validation, and test, respectively. Hyperparameters are tuned only on the validation set. All utterances are padded with zero or cut into 6-second audio and further scaled into [0,1]. The audio sample rate is 16kHz, thus each utterance is represented by a 96000-dimensional vector.

\subsection{Network Training}
The WaveRNN and WaveCNN models are trained separately for each paralinguistic task. The network structure is fixed for all experiments; the only hyperparameter we tune during training is the learning rate. We conduct a binary search for the learning rate in the range [1e-5, 1e-2]. The maximum number of epochs is 200. The best model and learning rate is selected according to the performance on the validation set.

Interestingly, we find that the WaveRNN and WaveCNN models perform very similarly. In the gender recognition task, both models have an accuracy of 88\%; in the emotion recognition task, the WaveRNN has an accuracy of 84\%, while the WaveCNN has an accuracy of 85\%; in the speaker recognition task, the WaveRNN has an accuracy of 69\%, while the WaveCNN has an accuracy of 73\%. This indicates that the convolutional back-end layers can also process audio sequences well.

\begin{table*}[t]
\centering
\small
\caption{Results of subjective human hearing test.}
\label{tab:subjectTest}
\begin{tabular}{|l|c|c|c|}
\hline
                                  & \multicolumn{1}{l|}{Emotion Recognition Accuracy} & \multicolumn{1}{l|}{Gender Recognition Accuracy} & \multicolumn{1}{l|}{Naturality} \\ \hline
Proposed Adversarial Examples     & 100\%                                             & 98\%                                             & 100\%                           \\ \hline
MFCC-based Reconstructed Examples & 84\%                                              & 58\%                                             & 6\%                             \\ \hline
White-Noise Added Examples       & 100\%                                             & 100\%                                            & 100\%                           \\ \hline
\end{tabular}
\end{table*}

\subsection{Adversarial Attack Evaluation}

In this work, we aim to attack WaveRNN, a state-of-the-art and widely used model for speech paralinguistic applications. However, due to the reasons mentioned in the previous section, it is not appropriate to set WaveRNN as the hypothetical model in the attack and we therefore use WaveCNN as our hypothetical model and expect that the attack can be generalized to the WaveRNN model. We perform the attack using FGSM described in Equation~\ref{equ:FGSM} on all three paralinguistic tasks with different perturbation factors $\epsilon$ and apply it twice (i.e., basic iterative FGSM~\cite{kurakin2016adversarial}). We can observe the following from the results shown in Figure~\ref{fig:attackResult}:
\begin{itemize}
    \item The proposed attack approach is effective in all paralinguistic tasks. With a perturbation factor of 0.02, the gender recognition error rate increases from 12\% to 31\%, and from 12\% to 25\% for WaveRNN and WaveCNN, respectively. With a perturbation factor of 0.015, the emotion recognition error rate increases from 16\% to 48\%, and from 15\% to 42.5\% for WaveRNN and WaveCNN,  respectively. With a perturbation factor of 0.032, the speaker recognition error rate increases from 31\% to 62\%, and from 29\% to 44\% for WaveRNN and WaveCNN, respectively. Note that 50\% and 75\% are the upper bound error rates for 2-class and 4-class classification tasks. 
    \item The proposed attack approach can be generalized, e.g., while the adversarial examples are designed to attack the WaveCNN as the hypothetical model, they can also cause a significant performance drop for the WaveRNN model. Note that the attack affects WaveCNN and WaveRNN in different ways; the error rate of the WaveCNN model increases linearly with respect to the perturbation factor, while the error rate of the WaveRNN model does not change with small perturbation factors, but changes rapidly when the perturbation factor exceeds a specific level. We observe a performance fluctuation of the WaveRNN attack in the gender recognition task, which we believe is due to the loss function coincidentally reaching a sub-optimal solution with the perturbation rather than a flaw of the attack. The error rate still increases with the perturbation factor after the fluctuation.
    \item The performance when there is no attack is not an indicator of the vulnerability of the model. In our experiments, the error rate of the emotion recognition model is similar to the gender recognition model when there is no attack, but increases much faster with larger perturbation factors than the gender recognition model. This indicates that even high performing models can be very vulnerable to adversarial examples. 
    
\end{itemize}

\subsection{Perturbation Analysis}

Our goal is to successfully fool the machine learning model, while also keeping the perturbation to be so small that it cannot be detected by a human. Toward this end, an adversarial example (generated for the emotion recognition task with $\epsilon=0.02$) is compared to an original example in Figure~\ref{fig:perturbation}. When $\epsilon = 0.02$, both the WaveRNN and WaveCNN models are close to random guesses with an error rate of 51\% and 46.5\%, respectively. Comparing the spectrograms of the proposed adversarial example and the adversarial example generated by the MFCC feature-level attack (i.e., Figure 1 in~\cite{itergenerating}), we observe that the proposed perturbation is much smaller, especially in the vocal parts. The feature-level attack greatly obscures the vocal spectrogram, while the proposed attack barely changes it. Quantitatively, we reproduced the MFCC based attacks in~\cite{itergenerating} and measured its minimal distortion (i.e., only the distortion caused by the audio reconstruction) is being equivalent to the distortion caused by our proposed perturbation when $\epsilon = 0.101$. Note that this is only the initial distortion level of the work in~\cite{itergenerating} (i.e., distortion due to the use of features) and no adversarial perturbation has been added yet. In our approach, where $\epsilon = 0.08<0.101$, all classifiers already predict like random guesses and the perturbation of the proposed approach is smaller than that of MFCC feature-level approaches for successful attacks.

In addition, we also performed a subjective human hearing test to evaluate the perturbation with respect to human perception. In this test, 10 subjects independent from this research were asked to listen to 5 proposed adversarial examples ($\epsilon=0.02$), 5 examples with equivalent levels of random white noise added, and 5 examples reconstructed from MFCC features as in~\cite{itergenerating} (i.e., no adversarial perturbation has been added and only the reconstruction distortion is present in the examples). The subjects are then asked to determine the gender, emotion, and naturality of each audio example. The speaker identification task was not included in this test, because the human subjects would need prior knowledge to identify the speaker. As shown in Table~\ref{tab:subjectTest}, 100\% and 98\% of the proposed adversarial samples are classified correctly in terms of emotion and gender, while 100\% of the proposed samples are also identified as natural speech. The results are 100\%, 100\%, and 100\% for white noise samples and 84\%, 58\%, and 6\% for MFCC reconstructed samples, respectively. This shows that the proposed adversarial examples are more comparable to natural speech than MFCC feature-based adversarial examples. In fact, listening to the proposed adversarial audio reveals that the vocal parts are unchanged, while the perturbations sound the same as ``normal'' noise. With such perturbations (i.e., added noise), humans have it difficult to detect an attack.

Finally, by comparing the spectrograms of the original example and the adversarial example in Figure~\ref{fig:perturbation}, we observe that the perturbation covers a broad spectrum, which means that it would be difficult to eliminate the attack through simple filtering.

\section{Discussion}

Since WaveCNN and WaveRNN have completely different back-end layers, we believe that one reason of the generalizability of adversarial examples between WaveCNN and WaveRNN is that they use the same front-end layers, which have a high probability of learning similar representations and therefore having similar convolutional kernel weights. Considering Equation~\ref{equ:activation}, the activation change of neurons is still very large. If this assumption is correct, then a variety of end-to-end speech processing models that use similar front-end layer structures might also suffer from the same attack.

More generally, the phenomenon that deep neural network based end-to-end speech paralinguistic models are robust to variance and noise in naturally occurring data, but vulnerable to man-made perturbation is somewhat counter-intuitive, but actually not surprising. End-to-end models experience performance improvements by processing problems in a much higher dimensional space in order to obtain a more accurate approximation function of the problem, but this also leaves blind spots in the space where data is distributed sparsely. Therefore, when manual perturbed data enters these blind spots, the model is not able to make correct predictions. Nonetheless, the problem is not unsolvable. If the exact type of attack is known, it can help with building a defense model by mixing adversarial examples into the training data to make the deep neural network see such adversarial examples and refine its approximation functions so that they can withstand adversarial examples~\cite{add1kurakin2016adversarial,add2madry2017towards,itergenerating}. To the best of our knowledge, current end-to-end audio processing algorithms (not just limited to speech paralinguistics) barely pay attention to this type of risk. Therefore, our goal is to provide a better understanding of such attacks to help design adversarial-robust models in the future. 

\section{Conclusions}

Adversarial attacks on computational paralinguistic systems pose a critical security risk that has not yet received the attention it deserves. In this work, we propose an end-to-end adversarial example generation scheme, which directly perturbs the raw audio waveform. Our experiments with three different paralinguistic tasks empirically show that the proposed approach can effectively attack WaveRNN models, a state-of-the-art deep neural network approach that is widely used in paralinguistic applications, while the added perturbation is much smaller compared to previous feature-based audio adversarial example generation techniques. 

\section{Appendix}
\subsection{Details of WaveRNN and WaveCNN}

The details of the network architectures for WaveCNN and WaveRNN are shown in Table~\ref{tab:arc}. The model training uses the following: Adam optimizer~\cite{kingma2014adam}, cross entropy loss function, learning rate decay of 0.1, max number of epochs of 200, and batch size of 100. The initial learning rate is selected within the range of [1e-5,1e-2] using a binary search.

\begin{table}[h]
\centering
\small
\caption{Network architecture of WaveCNN and WaveRNN}
\label{tab:arc}
\begin{tabular}{|p{1.1cm}|p{6cm}|}
\hline
Layer Index & \multicolumn{1}{c|}{Network Parameters and Explanation}                                                            \\ \hline  \hline
\multicolumn{2}{|c|}{Common Front-end Layers}                                                                                     \\ \hline
Reshape     & Divide audio into frames of 40ms                                                                                     \\ \hline
1-16        & 8$\times$(convolutional layers + max pooling layer), kernel size={[}1,40{]}, feature number=32, pool size=(1,2), zero padding \\ \hline  \hline
\multicolumn{2}{|c|}{WaveCNN Back-end Layers}                                                                                     \\ \hline
Reshape     & Concatenate output of each frame                                                                                    \\ \hline
17-28       & 6$\times$(convolutional layers + max pooling layer), kernel size={[}1,40{]}, feature number=32, pool size=(1,2), zero padding \\ \hline
Reshape     & Flatten                                                                                                             \\ \hline
29          & Fully-connected layer with 64 units                                                                                 \\ \hline
30          & Fully-connected layer with softmax output                                                                           \\ \hline  \hline
\multicolumn{2}{|c|}{WaveRNN Back-end Layers}                                                                                     \\ \hline
17          & LSTM recurrent layer with 64 units                                                                         \\ \hline
18          & Fully-connected layer with softmax output                                                                           \\ \hline
\end{tabular}
\end{table}

\bibliographystyle{ACM-Reference-Format}
\bibliography{main}

\end{document}